\documentclass[wcp]{jmlr}

\usepackage{longtable}
\usepackage{multirow}
\usepackage{makecell}
\usepackage{bm}
\usepackage{xspace}
\def\eg{\emph{e.g.}\@\xspace}
\def\ie{\emph{i.e.}\@\xspace}
\def\etc{\emph{etc.}\@\xspace}

\usepackage{booktabs}

\makeatletter
\let\Ginclude@graphics\@org@Ginclude@graphics 
\makeatother

\jmlrvolume{189}
\jmlryear{2022}
\jmlrworkshop{ACML 2022}

\title{Pose Guided Human Image Synthesis with Partially Decoupled GAN}

\author{
	\Name{Jianhan Wu}
	\Email{wujianhan@mail.ustc.edu.cn}\\
	\addr Ping An Technology (Shenzhen) Co.,Ltd., University of Science and Technology of China
	\AND
	\Name{Jianzong Wang}\thanks{Corresponding author: Jianzong Wang, jzwang@188.com}
	\Email{jzwang@188.com}\\
	\addr Ping An Technology (Shenzhen) Co., Ltd.
	\AND
	\Name{Shijing Si}
	\Email{shijing.si@outlook.com}\\
	\addr Ping An Technology (Shenzhen) Co., Ltd., School of economics and finance, Shanghai International Studies University
	\AND
	\Name{Xiaoyang Qu}
	\Email{quxiaoy@gmail.com}\\
	\addr Ping An Technology (Shenzhen) Co., Ltd.
	\AND
	\Name{Jing Xiao}
	\Email{xiaojing661@pingan.com.cn}\\
	\addr Ping An Technology (Shenzhen) Co., Ltd.
}

\editors{Emtiyaz Khan and Mehmet G\"{o}nen}

\begin{document}

\maketitle

\begin{abstract}
Pose Guided Human Image Synthesis (PGHIS) is a challenging task of transforming a human image from the reference pose to a target pose while preserving its style. Most existing methods encode the texture of the whole reference human image into a latent space, and then utilize a decoder to synthesize the image texture of the target pose. However, it is difficult to recover the detailed texture of the whole human image. To alleviate this problem, we propose a method by decoupling the human body into several parts (\eg, hair, face, hands, feet, \etc) and then using each of these parts to guide the synthesis of a realistic image of the person, which preserves the detailed information of the generated images. In addition, we design a multi-head attention-based module for PGHIS. Because most convolutional neural network-based methods have difficulty in modeling long-range dependency due to the convolutional operation, the long-range modeling capability of attention mechanism is more suitable than convolutional neural networks for pose transfer task, especially for sharp pose deformation. Extensive experiments on Market-1501 and DeepFashion datasets reveal that our method almost outperforms other existing state-of-the-art methods in terms of both qualitative and quantitative metrics. 
	
\end{abstract}

\section{Introduction}
Due to its portability and simplicity, two-dimensional (2D) human synthesis has become more and more popular in the multimedia and computer vision fields. Human image synthesis has huge potential in many commercial applications, such as virtual clothes try-on \cite{ren2021cloth}, person re-identification (Re-ID) \cite{zheng2015scalable} and so on. A challenging task of 2D human synthesis is pose guided human image synthesis (PGHIS) \cite{zhang2022exploring}, \cite{ren2022neural}, which renders the photo-realistic image of arbitrary pose from a reference pose human image. Because of the considerable changes in geometry and texture, rendering adequate details during pose transfer is challenging. 

Early image-based \cite{PG2} pose transfer methods directly combined human images, pose key points \cite{cao2017realtime}, and the inshop clothing representations together without taking into account the correlation between them, which brings missing and blurred texture. Most of existing methods process the pose and texture separately \cite{XingGAN}, \cite{zhang2021pise}, \cite{cui2021dressing}, \ie, two different encoders are used to map the reference pose and the whole human texture to the high-dimensional latent space, and then a decoder is used to fuse them to generate the human image of the target pose. This allows for generating human images with complete structure, but usually does not produce very realistic images with detailed information. Because the quality of synthesized images heavily depends on the encoder model, which is inherently challenging due to the difficulty of learning the topologically complex human texture \cite{zhang2022exploring}. Some methods make the encoder learn more details of human texture mainly by increasing the size of the model \cite{PG2}, \cite{SPIG}, but this improves the performance of detail rendering much less than the increasement of computational cost. Inspired by this, we argue that an encoder will learn more details when it encodes a part of a human body, so we propose to structurally decouple the human image. Specifically, we decouple the human image into 8 parts, then encode these parts using a weight-sharing encoder, and finally fuse them into a texture code to guide the synthesis of a realistic human image. This not only allows the encoder to focus on enough detailed texture information, but also ensures the lightweight nature of our model because of the weight-sharing operation.


In addition, due to the inherent structure of vanilla convolutional neural networks (CNN) \cite{zhao2022nnspeech} processing one local neighborhood at a time, they can only focus on short-range information, which results in their inability to achieve efficient spatial transformation \cite{transformer}. And CNN-based methods tend to have poor results when the pose transfer process involves large changes \cite{DIST}. Attention mechanism \cite{touvron2020deit}, \cite{zhang2019self}
computes the response of a target position as a weighted sum of all reference features, which let attention-based methods have the ability of long-range modeling. But in the pose transfer task, most of the existing methods either learn the correlation of the feature from different poses based on CNN or copy vanilla self-attention directly into image features. Few methods design a transformer module specifically for the pose transfer task. To fill this gap, we designed an efficient attention-based module (called transformer module) for pose transfer task. Specifically, the transformer module consists of two attention modules, the first one is the multi-headed self-attention module, which is responsible for capturing the correlation between the reference pose and the texture. The second is the multi-headed cross-attention module, which is used to migrate the reference texture features to the target pose. It is worthy that we introduce a Residual Fast Fourier Transform (Res FFT) \cite{mao2021deep} block instead of a traditional residual block into the transformer module, which makes the model focus on more detailed information like time-domain features and frequency-domain features \cite{wu2022improving}.

\begin{figure}
	\centering
	\includegraphics[width=0.9\linewidth]{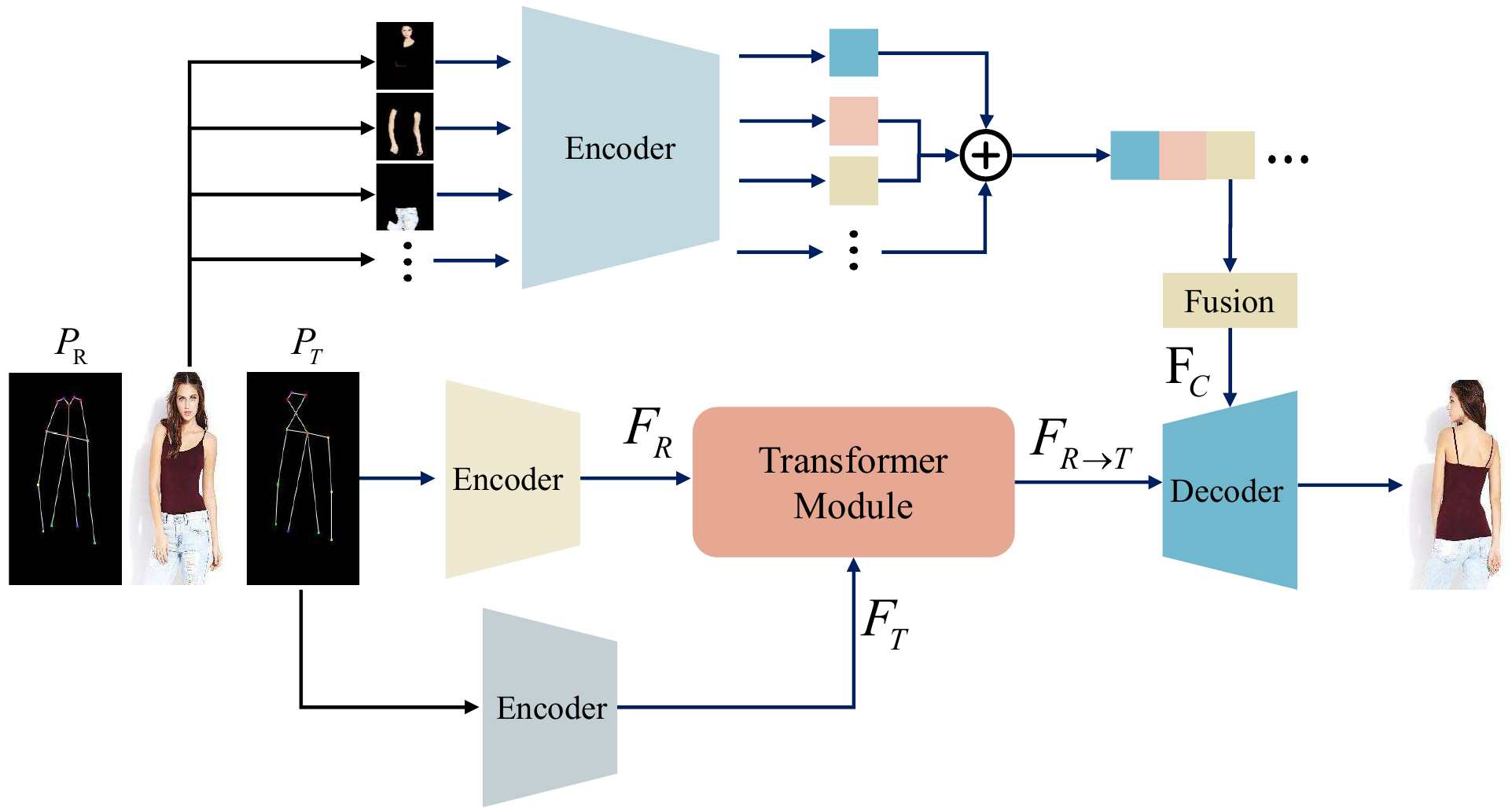}	
	\caption{Our generator mainly consists of a transformer branch and a body part decoupling branch. First, the reference pose, target pose, and reference image are concatenated together and mapped into a feature $ F_R$ by an encoder, and the target pose is mapped into a feature $ F_T$ by another encoder, and then the two features are passed through the transformer module to obtain the coarse feature $F_{R \rightarrow T} $ on target pose. The body part decoupling branch uses parser map to divide the person into different attributes and then encodes these parts into a series of style features using a global texture encoder. These style features are fused into a feature $ F_C$ by a covolutional layer, and the $ F_C$ is injected into the rough features of the target pose by controlling the affine transformation parameters in the AdaIN \cite{huang2017arbitrary} layer, and are finally decoded to a realistic image of target pose by the decoder.}
	\label{fig1}
\end{figure}

Based on the above ideas, we propose a Partially Decoupled GAN (PD-GAN) for PGHIS. The structure can be seen in Figure \ref{fig1}. Specifically, The body part decoupling branch and the feature texture transformer branch make up the two main branches of our model. The body part decoupling branch encodes the body parts into high-dimensional latent codes, which are then fused into features $ F_C$ that assist in generating realistic texture images. The transformer branch mainly renders the textures of the reference pose to the target pose, and we particularly design a new transformer module for the task. Our contribution can be summarized in the following three points:
\begin{itemize}
	\item We propose an effective model named PD-GAN for pose transfer task, which consists of a human partial decoupled module and transformer module. The two modules play crucial roles for generating realistic human images.
	\item To improve the ability of capturing the long-range dependency of pose transfer, we design a transformer module to obtain more global information. Besides, we use the Res FFT block that can capture both time-domain and frequency-domain information to replace the traditional residual block, which brings more realistic human images generated by our method.
	\item We perform extensive experiments to confirm the effectiveness of our method in comparison with other baseline methods. Additionally, a comprehensive ablation research indicates the role played by each component in the increased efficacy. 
\end{itemize}
\section{Related Work}

\textbf{Pose guided human image synthesis.} PGHIS was introduced by \cite{PG2}, which concatenates the reference pose, reference image and target pose together, then uses a coarse-to-fine network to generate the target pose image with the reference texture. This offen leads to texture misalignment. Some flow-based methods \cite{tabejamaat2021guided}, \cite{DIAF} obtain deeper correlations by estimating the correspondence between the flow of the pose and the appearance, which can alleviate the misalignment problem. However, the model size of such methods is too large and the training time is too long, which limits their applications. Some methods use a model framework with two branches controlling image texture and pose separately to solve the reference image and target image texture misalignment problem. For example, \cite{XingGAN} uses two blocks to model appearance and pose separately, \cite{VUnet} disentangle appearance and pose by combining the Autoencoder \cite{asru2021tang} and U-net \cite{ronneberger2015u}. In addition, to enhance the realism of the generated images, there are methods introduce parse map to assist the generator. PISE \cite{zhang2021pise} uses a network to generate a parse mapping of the target pose based on the parse mapping of the reference image and the target pose, and then uses this target pose mapping as an input to another network to assist in generating a target pose image with the texture of the reference image. These methods can mitigate the image texture misalignment problem well, but the extracted parse mapping maps from these methods are often unreliable and the pixel-level semantic annotations of datasets are very difficult to obtain, which limits their applications. 

All of these methods mentioned above encode the whole human image with an encoder, and there are few papers that decouple the person to focus on the human body parts. Unlike them, we use encoders to learn each part of the human body to improve the model's capacity to focus on details and thus enhancing PGHIS performance.

\vspace{0.1cm}
\noindent\textbf{Transformers in image synthesis.} The transformer was first introduced and widely used in NLP because it can compute in parallel and focus on global information. Many researchers have also applied it to the CV field and achieved good results in many tasks, such as image classification \cite{touvron2020deit}, \cite{wu2022augmentation}, detection \cite{deformdetr} and generation \cite{hudson2021ganformer}. For generation tasks, most transformer-based GAN models are designed for unconditional generation tasks, which are not suitable to conditional generation tasks (\eg, PGHIS tasks). The transformer is inherently superior to the CNN for PSGIS that requires attention to long-range dependencies, and existing methods only use the vanilla attention mechanism for PSGIS \cite{PATN}. Inspired by this, we specifically designed the transformer module to model complex topological structure for PSGIS.
\section{Method}
The framework of our end-to-end generator is depicted in Figure \ref{fig1}, which primarily consists of the transformer module and the body part decoupling module. The transformer module consists of some attention module, which adapts the long-range modeling capability of the transformer to alleviate the difficulties of the recovery of sharp poses. The body part decoupling module serves the decoder with Adaptive instance normalization (AdaIN) \cite{huang2017arbitrary} as an auxiliary role. Next, we will introduce the detailed design of every component of our model.
\subsection{Body Part Decoupling Module}
Most existing models process the whole body part through an encoder to obtain the body texture features, and the texture features obtained in this way tend to ignore some detailed information. To deal with this limitation, we introduce body part decoupling, an operation inspired by an intuitive idea: it is more difficult for an encoder to acquire the texture of all parts of the human body at once, so focusing on one small part of the human body at a time will allow the encoder to learn more detailed information. To implement this idea, we introduce the body part decoupling module, 
which first divides the human body into 8 parts using mask (respectively 'Hair', 'Upper Clothes', 'Dress', 'Pants', 'Face', 'Upper skin', 'leg' and ' background') \cite{men2020controllable}. Multiplying the reference human image with the component mask yields the decoupling human image with component i:
\begin{equation}
I_R^i = I_R \odot M_i
\end{equation}
Where the $ \odot$ and $M_i$ respectively are element-wise product and part masks. These component features are extracted by using weight-sharing encoders, and then concatenate these texture features. Finally the fused texture features are regarded as keys of the transformer module to generate a detailed image of the person. The fusion operation here is a 1$\times$1 convolution operation, the purpose of this operation is to select the appropriate body part texture to further assist in refining the generated human image.

\subsection{Transformer Module}
\begin{figure*}
	\centering
	\includegraphics[width=1.0\linewidth]{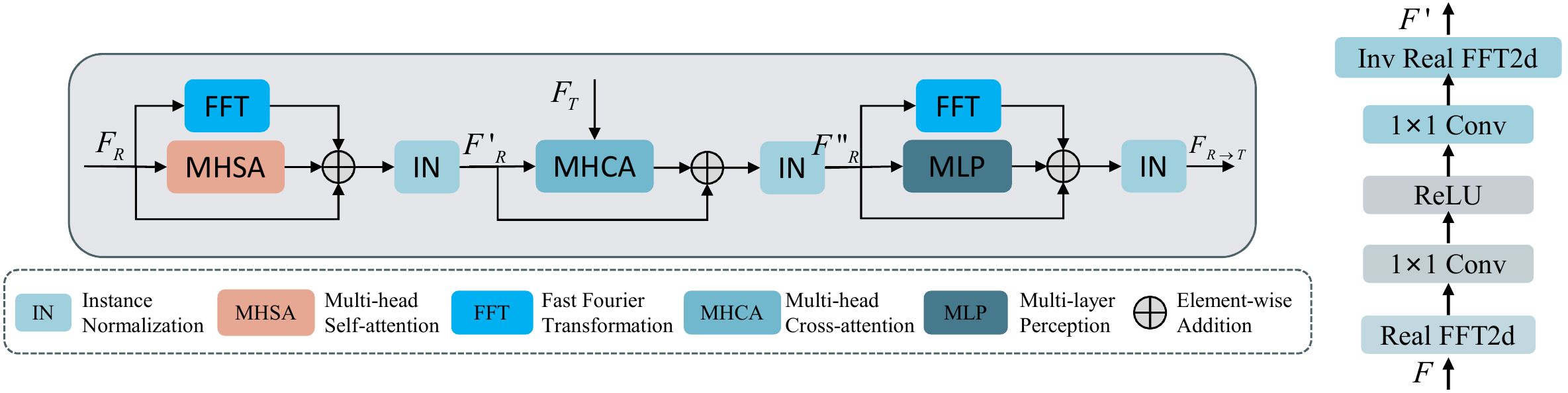}	
	\caption{The structure of transformer module. It consists of two attention-based blocks: MHSA and MHCA. MHSA is responsible for establishing the correlation between pose and texture, and MHCA is responsible for rendering the reference texture into the target pose.}
	\label{fig1-2}
\end{figure*}

Since the vanilla CNN-based transfer module is difficult to handle complex spatial deformations, we have designed a transformer module specifically for action transfer. As shown on the left of Figure \ref{fig1-2}, the transformer module contains two attention blocks: Multi-head Self-attention (MHSA) and Multi-head Cross-attention (MHCA). These blocks are built upon the Multi-head Attention (MHA), MHA is briefly defined as follows:
\begin{equation}
Attention(Q,K,V) = softmax(QK^T/\sqrt{d_k})
\end{equation}
\begin{equation}
head_i = Attention(QW^i_Q,KW^i_K,VW^i_V)
\end{equation}
\begin{equation}
MHA(Q,K,V) = concat(head_1, head_2,..., head_t)
\end{equation}
where Q, K, V denote query, key, value. $W^i_Q$, $W^i_K$, $W^i_V$ denote the learnable parameters of query, key and value, t denotes the number of attention heads. From this definition, it can be seen that the attention mechanism is computed over the whole feature variables, while it is fundamentally different from the local computation of CNN.

In our model, when Q=K=V, the MHA is used as MHSA. When Q=K, it is used as MHCA. Besides, unlike the traditional residual block, we use the Res FFT block here to focus on more detailed information. Because the attention mechanism performs well in focusing on global structural information, but its ability to focus on local detailed information is poor. FFT is a plug-and-play module that we found beneficial in the fields of image denoising and deblurring \cite{mao2021deep}. The structure of FFT is shown on the right side of Figure \ref{fig1-2}, the FFT process is as follows:
\begin{enumerate}
	\item performs Real FFT2d on F to obtain the frequency domain feature, where F $\in \mathbb{R}^{H*W*C} $, H, W, and C represent the height, width, and channel of the F.
	\item two convolution operations and an activation function are used to extract the local detail feature information, where the convolution operation and activation function respectively are two 1$\times$1 convolutional layers and an activation layer ReLU. 
	\item adopts inverse FFT2d operation to convert back to spatial (time-domain) features.
\end{enumerate}
where the Real FFT2d and Inv Real FFT2d operation is the same as the literature \cite{wu2022improving}. FFT block achieves attention to image details by combining frequency-domain information as residual connections with time-domain information. In summary, the transformer module can be formalized as follow:
\begin{equation}
	F'_R = IN( F_R + FFT(F_R) + MHSA(F_R,F_R,F_R))
\end{equation}
\begin{equation}
	F''_R = IN(F'_R + MHCA(F_T,F'_R,F'_R))
\end{equation}
\begin{equation}
	F_{R\rightarrow T}= IN(F''_R + MLP(F''_R) + FFT(F''_R))
\end{equation}
Where the IN is the operation of instance normalization, MLP denotes Multi-layer Perception. After N (2 in our experiment) time transformer modules, the final output is the feature $F_{R\rightarrow T}$.
\subsection{Loss Functions}
To generate realistic human images with arbitrary poses, several loss functions are employed to constrain the generator and discriminator to update their parameters. Since body parts were introduced to generate more realistic images, we designed two loss functions in our experiments, for the whole images and the partial components, respectively. 

\noindent\textbf{Loss functions for the whole images.} The loss consists of an L1, an adversarial, a perceptual and a style loss term, which can be formalized as follow:
\begin{equation}
	\mathcal{L}_{whole} = \lambda_1 \mathcal{L}_{\ell_1} +\lambda_2 \mathcal{L}_{adv} + \lambda_3 \mathcal{L}_{per} + \lambda_4 \mathcal{L}_{style} 
	\vspace{0cm}
\end{equation}
where $\lambda_1$, $\lambda_2$, $\lambda_3$ and $\lambda_4$ denote the weights of corresponding losses, respectively. The $\mathcal{L}_{\ell_1}$, $\mathcal{L}_{adv}$, $\mathcal{L}_{per}$, $\mathcal{L}_{style}$ can be defined as follow:
\begin{equation}
	\mathcal{L}_{\ell_1} = ||I_g - I_t||_1.
	\vspace{0cm}
\end{equation}
\begin{equation}
	\mathcal{L}_{adv} = \mathbb{E}[log(1-D(I_g))] + \mathbb{E}[logD(I_t)].
	\vspace{0cm}
\end{equation}
\begin{equation}
	\mathcal{L}_{per} = \sum_{i} ||\phi_i(I_t)-\phi_i(I_g)||_1.
	\vspace{0cm}
\end{equation}
\begin{equation}
	\mathcal{L}_{style} = \sum_{j} ||G^\phi_j(I_t)-G^\phi_j(I_g)||_1.
	\vspace{0cm}
\end{equation}
where $I_g$, $I_t$ stand the generated images by our model and the ground-truth images, $\mathbb{E}$ denotes the value of expectation, $\phi_i(*)$ denotes the feature of $[relui\_1]$ of the pre-trained VGG-19 model. $G^\phi_j$ represents the feature of $[relu2\_2, relu3\_4, relu4\_4, relu5\_2]$ of our generator. $D(*)$ represents the discriminator of our model, which is a multi-scale residual discriminator. Each of these loss functions has its role to play: By determining the pixel-level difference between the generated image and the ground truth image, the L1 loss controls the created image's overall effect. Adversarial loss with a discriminator helps the generator synthesize target pose images that resemble the texture of the reference image. Perceptual loss makes them as similar as possible at the deep feature level by mapping the generated image to the ground-truth image with a pre-trained VGG-19 network, which guides generating more ideal images at the human perceptual level. Style loss \cite{asru2021zhang} enables the generated image to be as feature-level same as feasible with the original image.

\noindent\textbf{Loss functions for the partial components.} Since we decouple the human body into multiple parts, aligning the generated images with the ground-truth images at the level of these parts can help generate more detailed human images. The partial loss function can be formalized as:
\begin{equation}
	\mathcal{L}_{par} = \sum_{j} ||\phi_j(C_i(I_t))-\phi_j(C_i(I_g))||_1.
	\vspace{0cm}
\end{equation}
Where $\phi_j$ denotes the j-th activation map of the pre-trained VGG-19 network, i denotes the i-part (\eg, face) cropping function according to the target poses. 

To sum up, the total loss of our image generator is:
\begin{equation}
	\mathcal{L}_{total} = \lambda_1 \mathcal{L}_{\ell} + \lambda_2 \mathcal{L}_{adv} + \lambda_3 \mathcal{L}_{per} + \lambda_4 \mathcal{L}_{style} + \lambda_5 \mathcal{L}_{par}
	\vspace{0cm}
\end{equation}
where $\lambda_1$, $\lambda_2$, $\lambda_3$ and $\lambda_4$ denote the weights of corresponding losses, respectively, which are the hyper-parameters of our model.

\section{Experiment Setup}
\subsection{Dataset}
We conduct experiments on DeepFashion \cite{liu2016deepfashion} dataset and Market 1501 \cite{zheng2015scalable} dataset. The DeepFashion dataset contains 52712 high-quality human models in different poses and clothes images, which are with a clean background. In our experiments, we split this dataset into 110426 pairs of images according to the processing of the PISE method. Among them, 101966 pairs of images are used for training and the others are used for testing. For comparison with other methods, we cropped all images to 256$\times$176. Market 1501 dataset contains 32668 images of campus people with complex background images of size 128$\times$64. We divide this dataset into 275632 pairs with the same setting as \cite{zhang2022exploring}, of which 263632 pairs are used as the training set and 12,000 pairs are used as the test set.

\subsection{Implementation Details}
The image size of DeepFashion (256$\times$176), Market-1501 (128$\times$64), batch size (64), the number of body part (8), total training iteration (400000) are the basic setup. The Adam optimizer is adpoted to train our model in our experiment. The learning rate (1$\times$$e-5$), head(h=2), number of transformer module (N=2), $\lambda_1$ (2), $\lambda_2$ (0.25), $\lambda_3$ (200), $\lambda_4$ (2.5) and $\lambda_5$ (0.5) are the network training hyperparameters (and the default settings we use). And we use 4 Tesla-V100 with 16G memory to experiment.

\subsection{Metrics}
To evaluate our generated images from different aspect, we employ three typical metrics for quantitative evaluation: peak signal-to-noise ratio (PSNR) \cite{hore2010image}, Frechet Inception Distance (FID)  \cite{heusel2017gans} , Learned Perceptual Image Patch Similarity (LPIPS) \cite{zhang2018unreasonable}. PSNR calculates the error between the generated images and the ground-truth images at the pixel value level, which discriminates the overall color accuracy of the generated image. FID calculates the difference at the distribution level, which discriminates the realism of the generated image. LPIPS calculates the difference at the feature level, which estimates the rationality of the generated image at the deep features.
\section{Experimental Results}
In this section, we perform extensive experiments to compare qualitatively and quantitatively with the state-of-the-art methods in recent years, and sufficient ablation experiments are performed to support the soundness of our model design.

\subsection{Qualitative Comparison}
\begin{figure}
	\centering
	\includegraphics[width=0.9\linewidth]{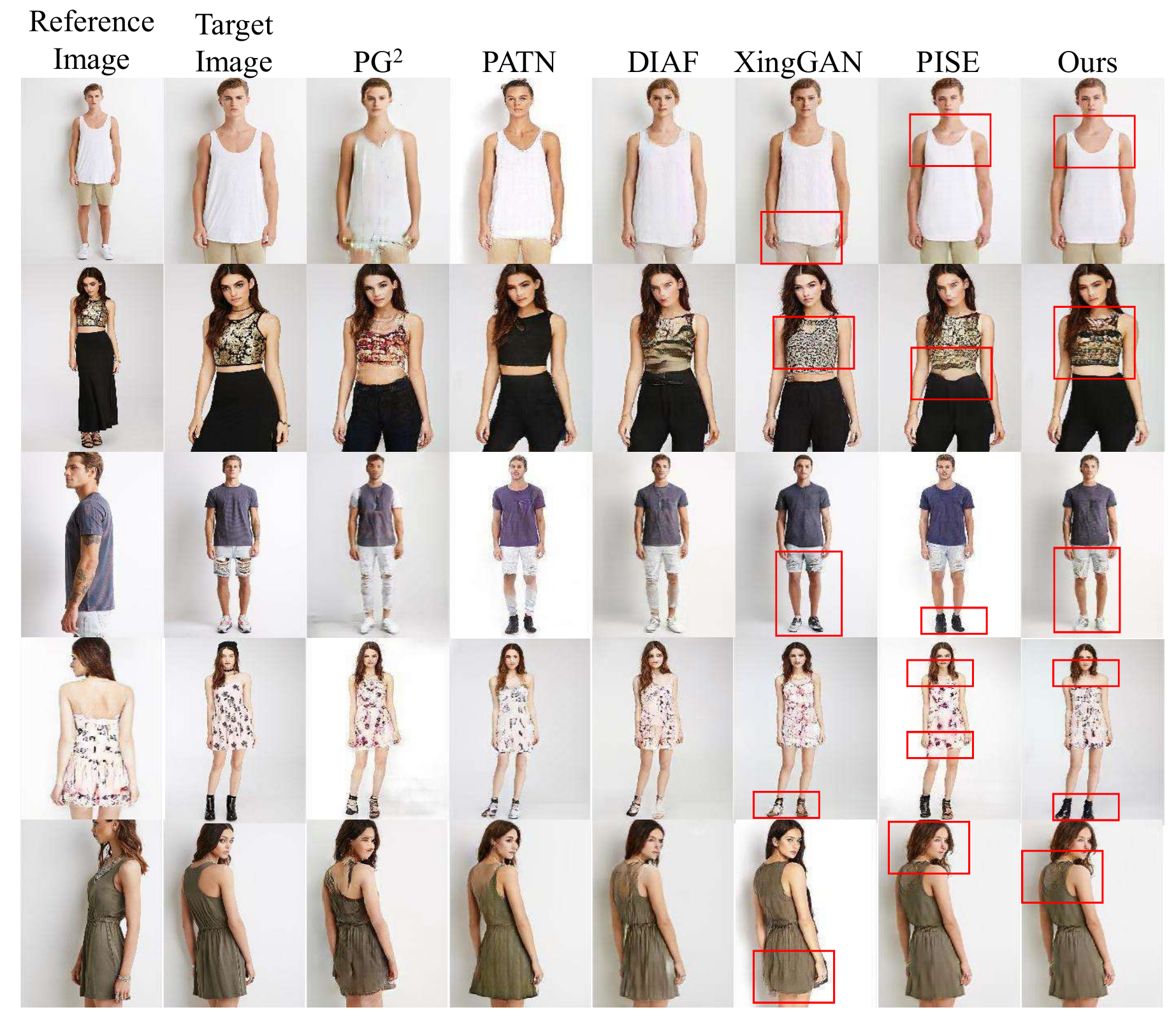}	
	\caption{Qualitative comparison results on DeepFashion dataset. Our method yields more realistic human images, especially in terms of detail. As you can see from the red box that is outlined in the figure.}
	\label{fig2}
\end{figure}

\begin{figure}
	\centering
	\includegraphics[width=0.7\linewidth, height=0.52\linewidth]{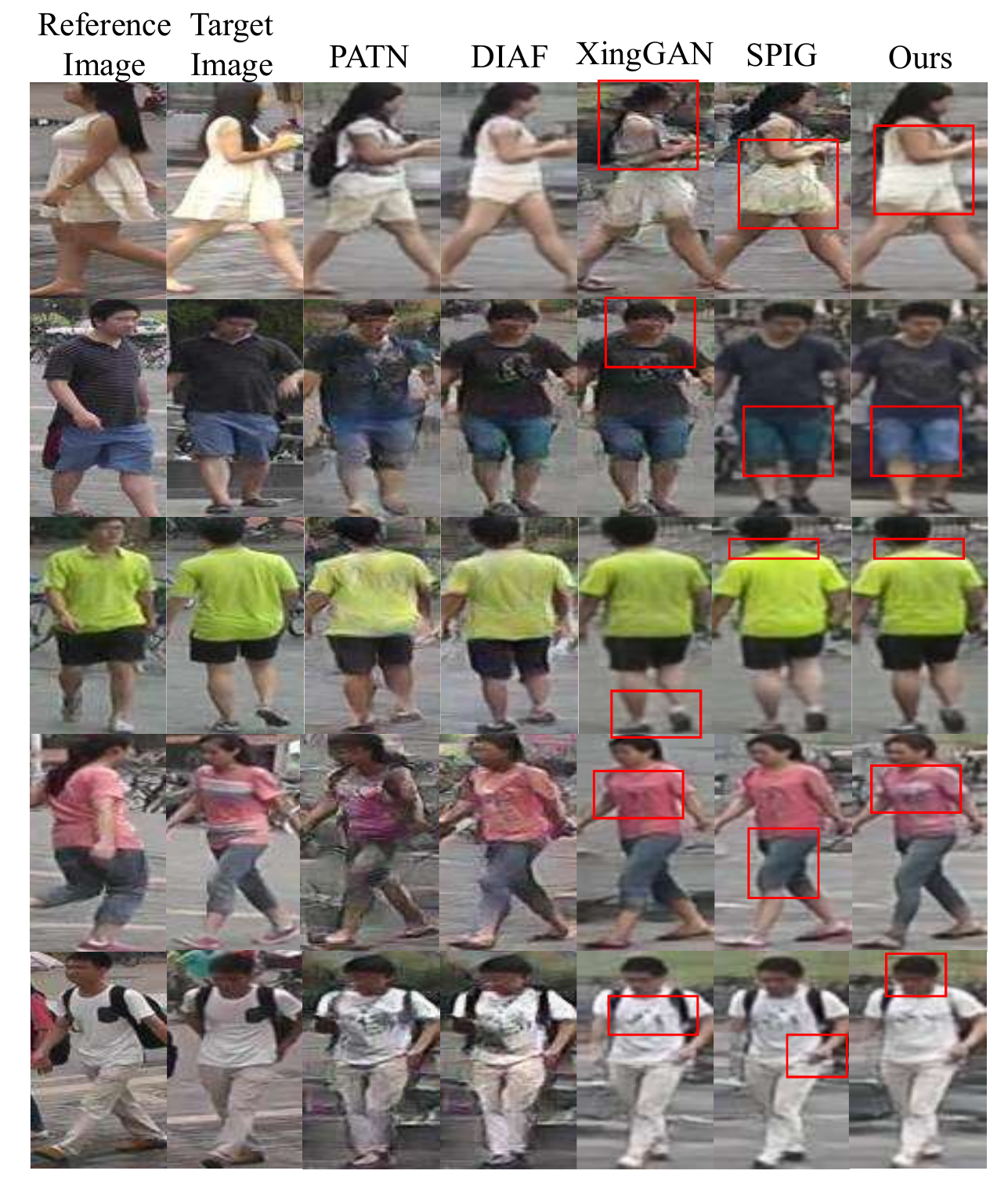}	
	\caption{Qualitative comparison results on Market1501 dataset. Although the dataset itself is not well trained, our method still produces better results compared to other baseline methods. As you can see from the red box that is outlined in the figure.}
	\label{fig3}
\end{figure}

As shown in Figures \ref{fig2} and \ref{fig3}, we have compared the human images generated by our method with several recent state-of-the-art methods on two datasets, these methods are: PG$^2$ \cite{PG2}, PATN \cite{PATN}, DIAF \cite{DIAF}, XingGAN \cite{XingGAN}, PISE \cite{zhang2021pise}, SPIG \cite{SPIG}, whose results were generated from the generated images and pre-trained models provided by the corresponding authors. For a fair comparison, we standardized the image size to 256$\times$176 for all methods on the DeeepFashion dataset and 128$\times$64 on Market-1501 dataset. The generated images by PISE and SPIG are better than the other baseline methods, including details such as clothes and colors. 
However, they are not as effective as they could be when transforming large movements (the fourth rows). Our method is designed to focus on more local detail information by decoupling the human body parts, and we have designed attention-based methods to handle the complex texture alignment. So our method shows superiority both in the overall rendering effect and in the local detail generation effect. Specifically, our method generates human images with more appropriate collar size and clothing color (the first rows and the second rows), which indicates that our method pays more attention to details. In addition, for large pose transfer (the third rows and the fifth rows), our method generates reasonable character textures for each compared to other methods, which reflects the role of attention mechanism in the overall rendering effect.

\subsection{Quantitative Comparison}
We compared the proposed method with several state-of-the-art methods including PG$^2$ \cite{PG2}, PATN \cite{PATN}, DIAF \cite{DIAF}, XingGAN \cite{XingGAN}, PISE \cite{zhang2021pise}, SPIG \cite{SPIG}.
To show fair competition, we uniformly crop the image size to 256$\times$176 for the DeepFashion dataset and 128$\times$64 for the Market-1501 dataset. 
And the evaluation results of these models are calculated based on their pre-trained models, the results are shown in Table \ref{tab:tab1}. The PISE and SPIG models gets more satisfied scores in three quantitative metrics than other baseline models, suggesting that the two-stage model and the region-adaptive normalization approach are at work. But it is obvious that our method gets the best score on the LPIPS metric on both two datasets, which indicates that the images generated by our method contain the most details and are the most consistent with human perception. And in PSNR and FID metrics, our method is also quite comparable with other models. For example, our method gets the best score on the PSNR metric on the DeepFashion dataset and the best on the FID metric on the Market-1501 dataset. Moreover, our model size is the smallest of any other models because we use the weight-sharing body part decoupling encoder, which is very important in application. 
\setlength{\tabcolsep}{0.83mm}{
	\begin{table*}
		\caption{Quantitative comparative results of image quality and model size with baseline methods. Our method achieves the best score for both PSNR and LPIPS on the DeepFashion dataset, while the FID are comparable than baseline methods. And on the Market-1501 dataset, the FID and LPIPS of our method are the best.}
		\renewcommand\arraystretch{1.2}
		\begin{tabular}{c|ccc|ccc|c}
			\Xhline{1pt}
			{\multirow{2}{*}{Model}} & \multicolumn{3}{c|}{DeepFashion}    & \multicolumn{3}{c|}{Market1501}      & \multirow{1}{*}{\begin{tabular}[c]{@{}c@{}}Model\\ Size $\downarrow$\end{tabular}} \\ \Xcline{2-7}{0.5pt}
			& PSNR $\uparrow$    & FID $\downarrow$     & LPIPS $\downarrow$  & PSNR $\uparrow$    & FID $\downarrow$    & LPIPS $\downarrow$ &                                                                                 \\ \hline
			PG$^2$ \cite{PG2}          & 17.53 & 49.56 & 0.2928  & 14.17 & 86.03  & 0.3619 & 437.09 M                                                                        \\
			DSC \cite{DSC}     & 18.09 & 21.27 & 0.2440 & 14.31 & 27.01  & 0.3029 & 82.08 M                                                                         \\
			PATN \cite{PATN}      & 18.25 & 20.75 & 0.2536  & 14.26 & 22.68  & 0.3194 & 41.36 M                                                                         \\
			DIAF \cite{DIAF}       & 16.90 & 14.88 & 0.2388  & 14.20 & 32.88  & 0.3059 & 49.58 M                                                                         \\                                                  
			XingGAN \cite{XingGAN}         & 17.92 & 39.32 & 0.2928 & 14.45 & 22.52 & 0.3058 & 42.77 M                                                                         \\
			PISE \cite{zhang2021pise}          & 18.52 & \textbf{11.51} & 0.2080  &   —       &    —       &    —     & 64.01 M                                                                         \\
			SPIG \cite{SPIG}         & 18.56 & 12.93 & 0.2128 & \textbf{14.47} & 23.11  & 0.2827 & 117.13 M                                                                         \\
			\hline
			\textbf{PD-GAN (Ours)}         & \textbf{18.64} & 13.13 & \textbf{0.2070}& 14.13 & \textbf{20.16} & \textbf{0.2827} & \textbf{18.52} M                                                                          \\
			\Xhline{1pt}
		\end{tabular}
		\label{tab:tab1}
	\end{table*}
}

\subsection{Ablation Study}
We train several ablation models to prove our hypotheses and validate the effect of our improvements. Below are the many alternatives for eliminating the corresponding components from the full model:

\vspace{0.2cm}
\noindent\textbf{The model without Transformer module (w/o Trans).} This model removes the Transformer module. In this way, the model becomes a pure CNN-based model, which lacks the ability of long-range dependency modeling.

\noindent\textbf{The model without Fast Fourier Transformation block (w/o FFT).} This model removes the FFT module from the transformer module, which means that the ability to focus on low frequency information is missing and the ability to focus on global information is weaken.

\noindent\textbf{The model without the part of face (pants, hair and so on) in the body decoupled part module (w/o Face (Pants, Hair)).} This model removes the face (pants, hair and so on) part in the body decoupled part module. 

\noindent\textbf{Full model (Full).} We use the PD-GAN model.

\begin{figure}
	\centering
	\includegraphics[width=0.78\linewidth, height=0.48\linewidth]{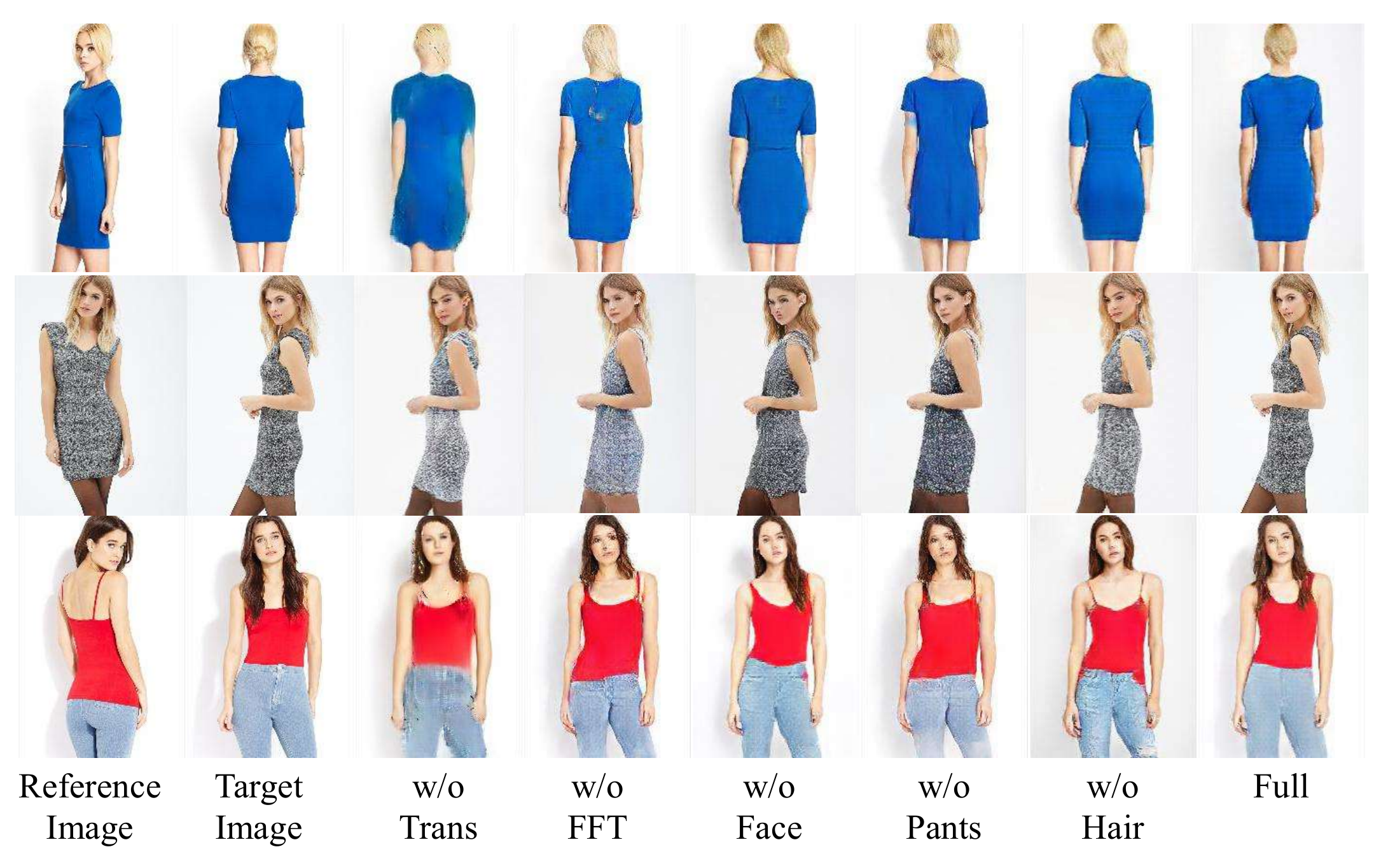}	
	\caption{Qualitative results of image quality on ablation study. Without transformer module has the greatest impact on the model to produce images, in addition, the absence of the FFT module also reduces the quality of the images. Also missing each part of the human body will have an impact on the clarity of the human image generation.}
	\label{fig5}
\end{figure}

\vspace{0.2cm}
The results on DeepFashion dataset are shown in Figure \ref{fig5} and Tabel \ref{tab2}. As shown in Figure \ref{fig5}, we can see that (1) w/o Trans method has the most severe results and fails to generate clear and complete images, which indicates transformer module plays an important role in building realistic human images. (2) Compare with w/o Trans, w/o FFT gets more sharp human images though gets worse results than the full model, which demonstrates FFT block with focusing on low frequency information can help generate detailed human images. (3) The quality of the images generated by w/o Face, w/o Pants and w/o Hair model is poor in the corresponding part than the full model, which means that each part of the human body part decoupled module will have an impact on the clarity of the human image generation. (4) As shown in Table \ref{tab2}, our Full model not only has satisfactory textural appearances but also gets the best scores of any other ablation model.

\subsection{Human Image Pose Transfer}
\begin{figure*}
	\centering
	\includegraphics[width=1.0\linewidth]{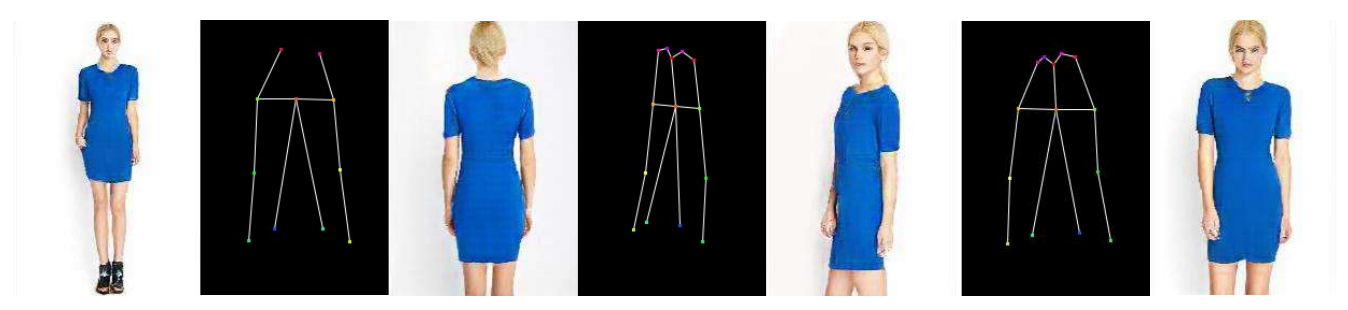}
	\caption{Our results of human image synthesis in different poses.}
	\label{fig4}
\end{figure*}

Human image pose transfer is to generate a target image with a target pose, given a reference image and a target pose. As shown in Figure \ref{fig4}, our model can generate realistic human images with details depending on different poses.
\setlength{\tabcolsep}{3.2mm}{
\begin{table*}[]
	\centering
	\caption{Quantitative results of image quality on ablation study. w/o Trans: without transformer module, w/o FFT: Transformer module without FFT block, w/o Face: lack of the part of face in the body decoupled part module, w/o Pants: lack of the part of pants in the body decoupled part module, w/o Hair: lack of the part of hair in the body decoupled part module, Full: our complete model.}
	\renewcommand\arraystretch{1.2}
	\begin{tabular}{l|l|l|l|l|l|l}
		\hline
		      & w/o Trans & w/o FFT & w/o Face & w/o Pants & w/o Hair & Full   \\ \hline
		PSNR $\uparrow$ & 13.39     & 13.45   & 13.67    & 13.79    & 13.88    & 14.13  \\ \hline
		FID  $\downarrow$ & 19.22     & 19.42   & 19.46    & 19.89    & 19.77    & 20.16  \\ \hline
		LPIPS $\downarrow$ & 0.3122    & 0.3021  & 0.3010   & 0.2991   & 0.2973   & 0.2827 \\ \hline
	\end{tabular}
	\label{tab2}
\end{table*}
}

\section{Conclusion and Discussion}
In this paper, we have proposed a novel model PD-GAN for PGHIS by explicitly controlling the pose of the reference human images. Specifically, we decouple the body parts to help the generation of realistic human images. For PGHIS, we design a new transformer module to handle human images with complex topology, which not only can capture the long-range dependency, but also focus on the details of human images using the FFT block. Extensive experiments show that the results of PD-GAN are qualitatively and quantitatively better than or competitive with existing state-of-the-art methods. Otherwise, the size of PD-GAN is the smallest, which is a big highlight in the application.

\begin{figure*}
	\centering
	\includegraphics[width=1.00\linewidth]{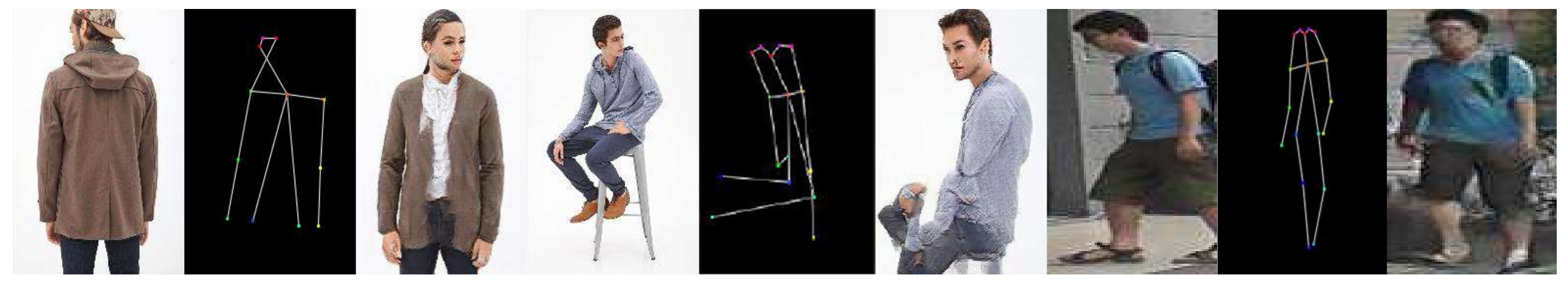}
	\caption{Failure cases. Our method generates less satisfactory images for very rare items in the dataset (\eg, stools), or when there are obvious interferences.}
	\label{fig6}
\end{figure*}

\vspace{0.1cm}
\noindent\textbf{Limitation.}Although our method produces satisfactory results in most cases, it still fails when the materials (\eg, the stool in \ref{fig6}) in the datasets are extremely lacking. We show some failure cases in Figure \ref{fig6}, inconsistencies and blur appeared in these images. We believe that training the model on a larger dataset would alleviate this situation where the person and object in images are confused. Furthermore, applying this model to more tasks (\eg, virtual try-on) is a future research direction for us, and we are committed to working on a generalized human image synthesis model.
\section{Acknowledgement}
This paper is supported by the Key Research and Development Program of Guangdong Province under grant No.2021B0101400003. Corresponding author is Jianzong Wang from Ping An Technology (Shenzhen) Co., Ltd (jzwang@188.com).
\bibliography{acml22}
\end{document}